\documentclass[11pt]{article}
\usepackage{paclic32}
\usepackage{times}
\usepackage{url}
\usepackage{latexsym}
\usepackage{amsmath}
\usepackage{multirow}
\usepackage{url}

\usepackage{graphicx}
\usepackage{booktabs}
\usepackage{subcaption}
\usepackage[whole]{bxcjkjatype}

\title{The Rule of Three: Abstractive Text Summarization in Three Bullet Points}

\author{
Tomonori Kodaira \\
  {\tt kdktmk@gmail.com} \\
  Graduate School of System Design \\
  Tokyo Metropolitan University \\
  \And
  Mamoru Komachi \\
  {\tt komachi@tmu.ac.jp} \\
  Graduate School of System Design \\
  Tokyo Metropolitan University \\
}

\date{}

\begin{document}
  \maketitle

  \begin{abstract}
    Neural network-based approaches have become widespread for abstractive text summarization.
    Though previously proposed models for abstractive text summarization
      addressed the problem of repetition of the same contents in the summary,
      they did not explicitly consider its information structure.
    One of the reasons these previous models failed to account for information
      structure in the generated summary is that
    standard datasets include summaries of variable lengths, resulting in
      problems in analyzing information flow, specifically, the manner in which
      the first sentence is related to the following sentences.
    Therefore, we use a dataset containing summaries with only three bullet points,
    and propose a neural network-based abstractive summarization model
      that considers the information structures of the generated summaries.
    Our experimental results show that the information structure of a summary can be controlled,
      thus improving the performance of the overall summarization.
  \end{abstract}

  \section{Introduction} \label{intro}
    Summarization can be achieved using two approaches, namely the extractive
    and abstractive approaches.
    The extractive approach involves selecting some part of a document (i.e.,
    sentence, phrase, or word) to construct a summary;
    in contrast, the abstractive approach involves generating a document
    summary with words that are not necessarily present in the document itself.

    Thus, the extractive approach \cite{Nallapati:16b,Li:13} is considered to
    yield a more grammatical summary than the abstractive approach,
    because the former involves directly extracting output expressions from the source text.
    However, as is obvious, because words that are not present in the source
    text cannot be selected in the case of the extractive approach,
    abstractive approaches are becoming increasingly popular for automatic
    summarization tasks.

    In previous work, \newcite{Rush:15} proposed an abstractive sentence
    summarization method that involves generating novel words
    in a summary based on the sequence-to-sequence model proposed by \newcite{Sutskever:14}.
    Furthermore, recently, improvements to the abstractive text summarization
    method were proposed by \cite{Nallapati:16a,See:17}.
    Although their proposed model generates fluent summaries owing to the use
    of a large-scale dataset, 
    it cannot produce a structured summarization, because it is trained using
    the CNN / Daily Mail datasets, which are not annotated with any structural information.

    Therefore, in this work, we focus on generating a structured summary for a
    document, in particular, a summary in three sentences.
    Because the CNN / Daily Mail datasets include summaries with a varying number of sentences,
    they cannot be annotated with information structures directly.
    Considering this, we employ a Japanese summarization dataset from Livedoor
    News, whose size is the same as the CNN / Daily Mail datasets.
    Because Livedoor News broadcasts news with a summary in three sentences,
    it is easy to analyze summaries using  this dataset.

    To produce a summary in three bullet points, 
    we first annotate the dataset with an information structure.
    Then, we train a binary classifier using the information structure of
    summaries to build two summarization sub-models on our dataset.
    Finally, the obtained summarization model selects the summary structure
    based on the input, and generates the summary according to the desired structure.

    The contributions of our work are as follows:

    \begin{itemize}
      \item We annotated and analyzed the structure of summaries in a Japanese
          news summarization dataset, whose summaries are in the form of three
            sentences.\footnote{\url{https://github.com/KodairaTomonori/ThreeLineSummaryDataset/}}
      \item Our proposed model generates a summary in three bullet points.
    \end{itemize}

  \section{Related Works} \label{related}
  \subsection{Dataset}
    In the case of abstractive sentence summarization, \newcite{Rush:15}
    proposed a new summarization method to generate an abstractive summary using a
    sequence-to-sequence model; in particular, they achieved state-of-the-art
    performance on the DUC-2004 and Gigaword corpora.

    In contrast, for abstractive text summarization using the CNN / Daily Mail datasets, 
    the objective is to output a summary of an article consisting of
    multi-sentences.
    \newcite{Nallapati:16a} proposed an improved summarization model for this task,
    which is essentially an attention encoder-decoder model, including a trick
    to use a large vocabulary \cite{Jean:15}, switching pointer-generator mechanism, and hierarchical networks.
    They proposed a new dataset for multi-sentence summarization and
    established a benchmark using this dataset.

    However, these works did not address the problem of information structure
    in a generated summary. Therefore, in our work, we attempt to consider the information
    structure of a summary to further improve the summarization model.

  \subsection{Model} \label{coverage_model}
    Currently, the model proposed by \newcite{See:17} is the state-of-the-art
    summarization model on the CNN / Daily Mail datasets.
    In particular, their baseline model is based on the one proposed by \cite{Nallapati:16a},
    which is further improved by including a hybrid pointer-generator network and coverage mechanism.
    Because our model is based on their model, we describe their model in
    greater detail in this subsection.


    \paragraph{Attention encoder decoder.} \label{coverage_model:seq2seq}
      Let the input sequences be tokens of the article $w_i$ and the output
    sequences be tokens of the summary $y_i$.
      A bidirectional long short-term memory (LSTM) network is used as the
    encoder, whereas a unidirectional LSTM is used as the decoder.
      A sequence of encoder hidden states $h_i$ are produced by the encoder. 
      At each step $t$, the decoder receives the word embedding of the previous word
      \footnote{During training, this is the previous word of the reference;
      however, during testing, this is the previous word output by the decoder.},
      and has decoder state $s_t$.
      The attention distribution $a^t$ is calculated as in \cite{Bahdanau:15}:
      \begin{eqnarray}
        e^t_i & = & v^T\tanh(W_hh_i + W_ss^t + b_{a})  \\
        a^t   & = & {\rm{softmax}}(e^t)
      \end{eqnarray}
      \noindent
      where $v, W_h, W_s$, and $ b_{a}$ are learnable parameters.
      The attention distribution indicates the importance of the encoder hidden states as a probability distribution at time step $t$.
      Furthermore, the context vector $h_t^*$ is computed as follows: 
      \begin{equation}
        h^*_t = \sum_i a^t_i h_i
      \end{equation}
      \noindent
      The context vector is concatenated with the decoder state $s_t$ and
        input through two linear layers to produce the vocabulary distribution $P_{vocab}$:
      \begin{equation}
        P_{vocab} =  {\rm{softmax}}(V'(V([s_t,h^*_t] + b) + b'))
      \end{equation}
      \noindent
      where $V, V', b$ and $b'$ are learnable parameters.
      While training, the loss at timestep $t$ is the negative log likelihood
      of the target word $w^*_t$ for that timestep, which is used
      to calculate the overall loss for the entire sequense:
      \begin{eqnarray}
        {\rm{loss}}_t &=& - \log P_{vocab}(w_t^*) \\
        {\rm{loss}}   &=& \frac{1}{T}\sum^T_{t=0} {\rm{loss}}_t
      \end{eqnarray}

    \paragraph{Hybrid pointer-generator network.} \label{coverage_model:pg}
      \newcite{See:17} also proposed a hybrid pointer-generator network, which
    combines the attention and vocabulary distributions.
      In particular, their pointer-generator network is a hybrid between a
    sequence-to-sequence attention model (Section \ref{coverage_model:seq2seq})
    and pointer network \cite{Vinyals:15}.
      Thus, this network considers the source as well as target word
    distributions, and thereby addresses the problem of unknown word generation.
      In the pointer-generator model, the generation probability $p_{gen} \in$
    [0, 1] at time step $t$ is calculated from the context vector $h_t^*$,
    decoder state $s_t$, and decoder input $x_t$:
      \begin{equation}
        p_{gen} = \sigma(w^T_{h^*}h^*_t + w^T_s s_t + w^T_x x_t + b_g)
      \end{equation}
      \noindent
      where vectors $w_{h^*}, w_s, w_x$, and scalar $b_{g}$ are learnable parameters,
      and $\sigma$ represents the sigmoid function. 
      $p_{gen}$ is used as a soft switch to select a word from the vocabulary
    distribution $P_{vocab}$ or a word from the attention distribution $a^t$.
      For each document, the authors developed an extended vocabulary, which
    is the union of the vocabulary and all words in the source document.
      The probability of the extended vocabulary is calculated as follows:
      \begin{equation}
        P(w) = p_{gen}P_{vocab}(w) + (1 - p_{gen}) \sum_{i:w_i=w}a^t_i
      \end{equation}
      If $w$ is an out-of-vocabulary (OOV) word, $P_{vocab}(w)$ is 0; in
    addition, if $w$ does not exist in the source words, $\sum_{i:w_i=w}a^t_i$ is 0.

    \paragraph{Coverage mechanism.} \label{coverage_model:cov}
      Aside from the above-mentioned changes, \newcite{See:17} improved the
    coverage model \cite{Tu:16} to address the repetition problem.
      In their model, a coverage vector $c^t$, which is the sum of attention
    distributions at all previous decoder timesteps, is saved:
      \begin{equation}
        c^t = \sum_{t'=0}^{t-1}a^{t'}
      \end{equation}
      \noindent
      where $c^t$ is a distribution over the source document words that
    indicates the magnitude of attention each word in the source document
    receives until timestep $t$.
      The coverage vector is applied to the attention mechanism as follows:
      \begin{equation}
        e^t_i = v^T\tanh(W_h h_i + W_s s_t + w_c c^t_i + b_{a})
      \end{equation} 
      \noindent
      where vector $w_c$ is a learnable parameter.
      This prevents the attention mechanism from repeatedly visiting the same
    location in the document, and thus avoids generating repetitive text.
      In addition, they constructed a new loss function to incorporate the
    coverage loss to penalize repeated attention to the same location.
      \begin{equation}
        {\rm{loss}}_t = - \log P(w_t^*) + \lambda \sum_i \min(a^t_i, c^t_i)
      \end{equation}

  \section{Annotation of the Summary Structure} \label{dataset}
    \subsection{Dataset} \label{dataset:char}
    We crawled pairs of Japanese articles and summaries from Livedoor News
    \footnote{http://news.livedoor.com/} in a manner similar to the previous
    work \cite{Tanaka:16}.
    These summaries are written by human editors.
    In particular, these summaries consist of exactly three sentences, which
    we will discuss in detail later.
    We crawled data from January 2014 to December 2016, which included 214,120
    pairs of articles and summaries.
    We divided these pairs into 211,744 training pairs, 1,200 development pairs, and 1,200 test pairs.
    The development and test pairs were extracted from the data of January
    2016 to December 2016, which included 100 pairs of articles and summaries per month.
    
      Each article was tagged with a category selected from the nine
    primary categories\footnote{National, World, IT Business, Entertainment,
    Sports, Movies, Foods, Lifestyle (Women), and Latest.} as well as a subcategory selected from the
    several available subcategories.
      Furthermore, the articles also included some special tags (such as
    keywords, key phrases, or more specific category information).
      In the crawled data, each news item includes a title and article, as
    well as a shorter title and abstractive summary.
      
      However, in our experiments, we only use the article and summary from the dataset.
      Other useful information will be exploited in future work.

      \begin{figure}[t]
        \center
        \includegraphics[width=0.8\columnwidth]{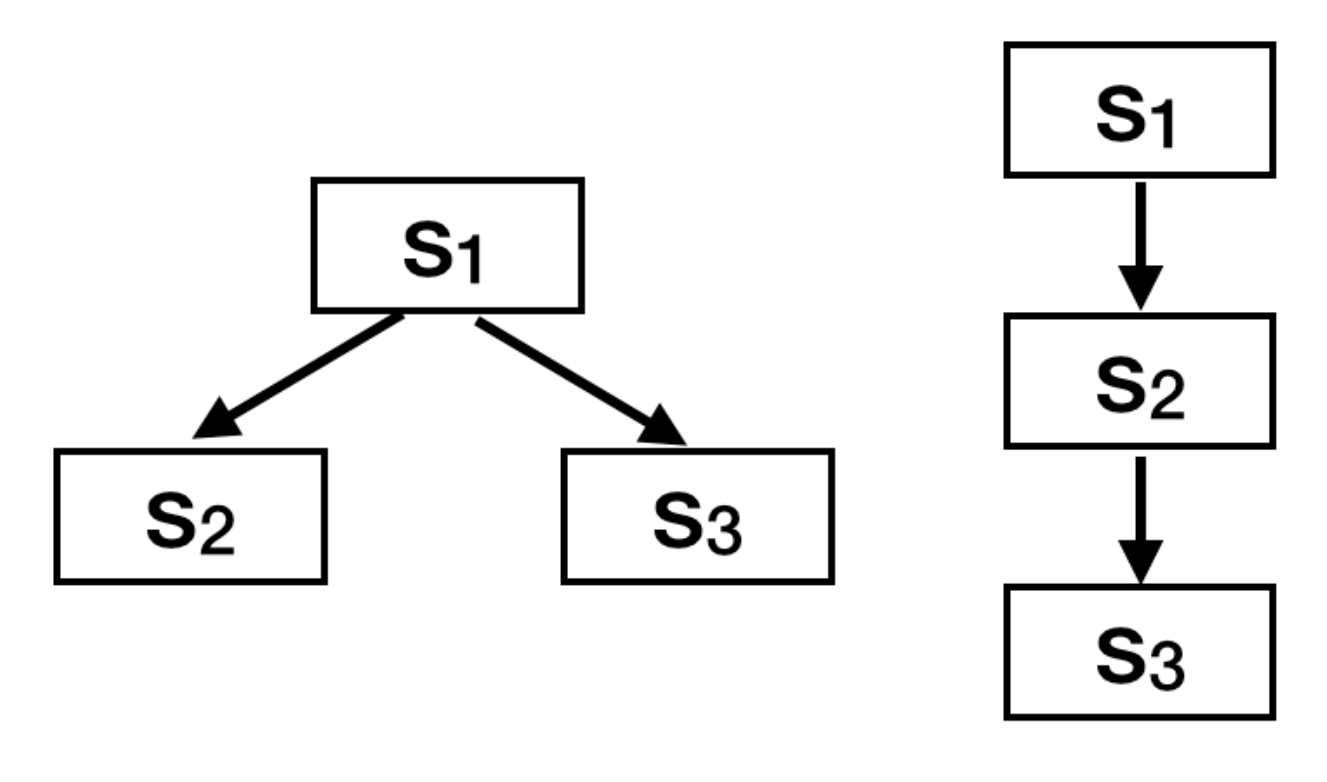}
          \caption{Parallel and sequence summary structures.}
        \label{ss}
      \end{figure}

    \subsection{Annotation} \label{dataset:anno}
      The summaries of the Livedoor News dataset consist of three sentences, which enables us to analyze the structure of the output.
      We annotated summaries with the structure of sentences for the development and test data.
      Summaries are typically single sentences whose lengths are less than
      36 characters (counted as full-width). The average length of the
      summaries in the test set is 32.5. We did not perform any pre-processing
      for annotation, and did not refer to the original article to assign
      labels.

      After a preliminary investigation, we divide the summaries into four types: \textit{parallel},
        \textit{parallel with enumeration}, \textit{sequence}, and
\textit{sequence with segmented sentences}.
      Enumeration in a summary is when recommended items are introduced,
      whereas segmented sentences in a summary are those that were originally part of a longer sentence
      in the article.

      However, almost all summaries are divided into the following two types:
      \textit{parallel} and \textit{sequence}.
      Figure \ref{ss} illustrates the difference between these types.
      In both these summary types, the first sentence describes the primary
      incident, and the second sentence contains additional information about the primary incident.
      Then, in the case of the parallel type, the third sentence explains the
    first sentence;
      however, its content is different from that of the second sentence. 
      In contrast, in the case of the sequence type, the third sentence
    includes detailed information about the second sentence.   
      Thus, the parallel types have no particular order in terms of the second
    and third sentences, whereas the sequence types have these two sequences
    in order.

      In addition, the subject is often omitted in Japanese, thus the third sentence's
        zero-subject is almost the same as the second sentence's subject.
      When we annotate such a sentence, the summary is marked as an instance
    that requires zero anaphora resolution to generate an appropriate output.

    \subsection{Analysis} \label{dataset:result_analysis}

      \begin{table}[t]
        \centering
        \begin{tabular}{lrrr}
        \toprule
                                        & dev  & test  & total \\ \midrule
          parallel                      & 843  & 755   & 1,598 \\
          parallel w/ enumeration       & 72   & 65    & 137   \\
          sequence                      & 268  & 275   & 543   \\
          sequence w/ segmented sents.  & 12   & 5     & 17    \\
        \bottomrule
        \end{tabular}
        \caption{Results of annotation of the summaries.}
        \label{anno_result}
      \end{table}

      Table \ref{anno_result} lists the results of annotation of the summaries
    of the Livedoor News dataset.
      Approximately 80\% of the summaries are tagged as {\it{parallel}}, while the
    remainder are tagged as {\it{sequence}}.
      In a {\it{sequence}} summary, the second and third sentences simply
    indicate examples related to the first sentence, but not in the form of a sentence.

      In addition, annotation revealed that the first sentence is similar to the title, and
        almost all the second sentences include additional information such as
        an example of the first sentence.
      Therefore, the first and second sentences can be successfully generated
      by existing models.

      However, the third sentence plays various roles in this dataset.
      In particular, in a \textit{sequence} summary, the third sentence is
    based on the second sentence, whereas in a \textit{parallel} summary, the
    third sentence is based on the first sentence.
      Thus, our proposed model uses this characteristic to generate the third sentence
    in a summary.

  \section{Structure-aware Summarization Model} \label{cls_model}
    \begin{figure}[t]
      \center
      \includegraphics[width=\columnwidth]{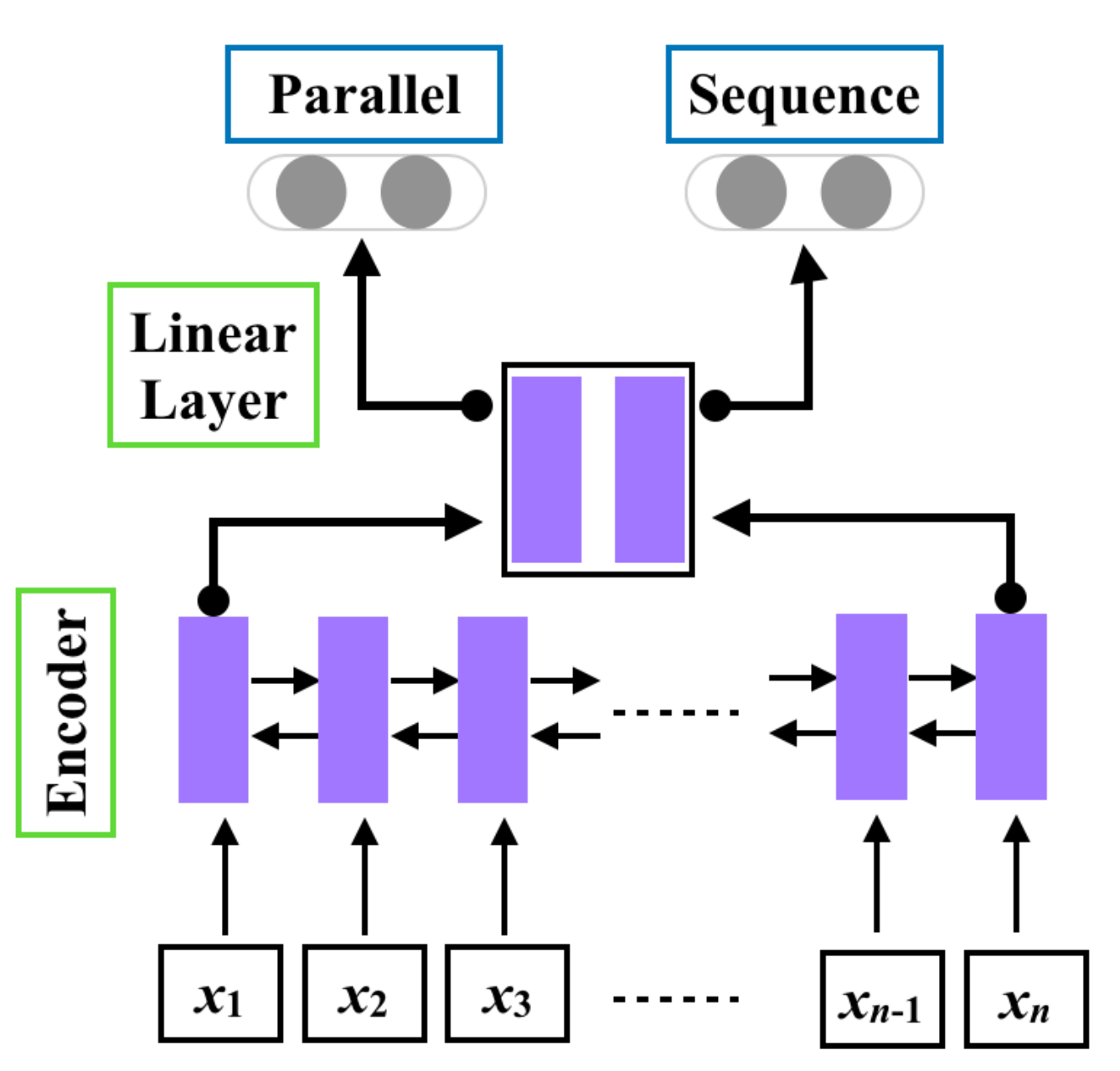}
      \caption{Summary structure classification.}
      \label{naive_model}
    \end{figure}

    We generate a summary considering its structure.
    First, we predict the structure of the summary to be generated,
    and then generate the summary according to the predicted structure.

    \paragraph{Summary structure classification.} \label{cls_model:model}
    First, we need to train structure-specific summarization models for the parallel and sequence
    types.  However, because the training dataset is not annotated with summary types,
    we build a binary classifier for summary types using summary information
    to label the summaries.

      Using the binary classifier, the summaries are assigned to either
    {\it{parallel}} or {\it{sequence}} types (Figure \ref{naive_model}).
      Let the input sequences be tokens of a summary $x_i$ and the output
    label be $l$.
      We use a bidirectional LSTM as an encoder.
      A sequence of encoder hidden states $h_i$ is produced by the encoder.
      Furthermore, the final hidden states of the forward and backward encoders
    are concatenated.
      Finally, a linear transformation is applied to the vector $h$ for each
      type:
      \begin {eqnarray}
        h &=& [h^{forward}_{n}, h^{backward}_{1}] \\
        y_{parallel} &=& {\mathrm{softmax}}(W_p h + b_p)  \\
        y_{sequence} &=& {\mathrm{softmax}}(W_s h + b_s)
      \end{eqnarray}
      \noindent
      where $W_p$, $W_s$, $b_p$ and $b_s$ are learnable parameters.

    \paragraph{Structure-specific sub-models.} \label{cls_model:finetune}
    Second, we construct the structure-specific summarization models using the automatically
    annotated dataset.

      In Section \ref{coverage_model}, we describe the base summarization model \cite{See:17}.
      %
      In particular, we pre-train the structure-specific summarization models using all
        training data as in \cite{See:17} regardless of the summary structures,
        and then perform fine-tuning for each type using the automatically
        annotated dataset.

    \paragraph{Structure-aware summarization.} \label{cls_model:select}
    Finally, we build another binary classifier for summary types; however, in
    this case, we use only articles, because summaries are not available
    in the test data.
    Then, we generate a summary based on the type predicted by the
    second summary-type classifier using the structure-specific summarization models.

      To decide which model should be used to generate a summary, we construct another summary
        structure classification model in a manner similar to the one
    specified above.
      In this case, because summaries cannot be used as input during testing,
    articles are used for training; based on the classification results of
    this model, our proposed summarization model selects a structure-specific
    summarization model to output the final summary.

  \section{Experiments} \label{exp_cls}

    \begin{table}[t]
      \center
      \begin{tabular}{cccc}
        \toprule
                    & precision & recall & F1 \\ \midrule
        parallel    & 0.897     & 0.325  & 0.477  \\
        sequence    & 0.833     & 0.156  & 0.263  \\
        \bottomrule
      \end{tabular}
      \caption{Classification results based on summaries.}
      \label{cls_result}

      \center
      \begin{tabular}{cccr}
        \toprule
                    & precision  & recall & F1   \\ \midrule
        parallel    & 0.71       &  0.53  & 0.61 \\
        sequence    & 0.25       &  0.42  & 0.31 \\
        \bottomrule
      \end{tabular}
      \caption{Classification results based on articles.}
      \label{cls_result2}
    \end{table}

    \begin{table*}[t]
  \center
  \scalebox{0.91}{
    \begin{tabular}{c|rrr|rrr|rrr|rrr}
      \hline
      & \multicolumn{3}{c|}{Coverage} & \multicolumn{3}{c|}{Parallel\_model}
        & \multicolumn{3}{c|}{Sequence\_model} & \multicolumn{3}{c}{Proposed} \\ \hline 
      & \multicolumn{1}{c}{R-1} & \multicolumn{1}{c}{R-2} & \multicolumn{1}{c|}{R-L} &
        \multicolumn{1}{c}{R-1} & \multicolumn{1}{c}{R-2} & \multicolumn{1}{c|}{R-L} &
        \multicolumn{1}{c}{R-1} & \multicolumn{1}{c}{R-2} & \multicolumn{1}{c|}{R-L} &
        \multicolumn{1}{c}{R-1} & \multicolumn{1}{c}{R-2} & \multicolumn{1}{c}{R-L} \\ \hline

      All      & 47.22 & 21.65 & 32.96 &
                 47.41 & 21.75 & 33.36 &
                 {\bf 47.88} & {\bf 21.92} & 33.40 & 
                 47.69 & 21.75 & {\bf 33.41} \\
      Parallel & 47.02 & 21.83 & 32.92 &
                 47.16 & 21.94 & 33.21 &
                 47.43 & 21.97 & 33.20 &
                 {\bf 47.83} & {\bf 22.00} & {\bf 33.63} \\
      Sequence & 47.80 & 21.20 & 33.05 &
                 48.14 & 21.34 & {\bf 33.85} & 
                 {\bf 49.11} & {\bf 21.86} & 33.80 &
                 47.34 & 21.09 & 32.80 \\
      \hline
    \end{tabular}
  }
  \caption{Evaluation results for three bullets summarization.
    `Coverage' is baseline model which is proposed by \newcite{See:17}.
    `Parallel\_model' and `Sequence\_model' are sub-models of the proposed
method.
    }
  \label{plain_result}
\end{table*}

    \begin{table*}[t]
  \center
  \scalebox{0.95}{
    \begin{tabular}{c|r@{ }r|rrr|rrr|rrr}
      \hline
      &&&\multicolumn{3}{c|}{1st} & \multicolumn{3}{c|}{2nd} & \multicolumn{3}{c}{3rd} \\ \hline 
      pair & \multicolumn{2}{c|}{\# of pairs} 
      & \multicolumn{1}{c}{R-1} & \multicolumn{1}{c}{R-2} & \multicolumn{1}{c|}{R-L} &
        \multicolumn{1}{c}{R-1} & \multicolumn{1}{c}{R-2} & \multicolumn{1}{c|}{R-L} &
        \multicolumn{1}{c}{R-1} & \multicolumn{1}{c}{R-2} & \multicolumn{1}{c}{R-L} \\ \hline
      123 & 528 &  (44.0\%) &
        55.07 & 34.56 & 40.22 &
        36.93 & 17.31 & 25.21 & 
        {\bf 32.64} & {\bf 13.19} & {\bf 22.62} \\ 
      132 & 381 &  (31.8\%) &
        52.38 & 31.94 & 37.72 &
        33.59 & 14.14 & 23.15 &
        {\bf 30.83} & {\bf 11.00} & {\bf 20.93} \\ 
      213 & 100 &  (8.3\%) &
        38.03 & 18.30 & 26.50 &
        {\bf 31.32} & {\bf 10.34} & {\bf 21.25} &
        34.73 & 15.47 & 24.64 \\ 
      231 & 66  &  (5.5\%) &
        41.46 & 21.95 & 28.09 &
        37.22 & 17.95 & 27.23 &
        {\bf 26.96} & {\bf 6.02} & {\bf 18.08} \\ 
      312 & 64  &  (5.3\%) & 
        {\bf 28.37} & {\bf 7.78} & {\bf 19.26} & 
        34.19 & 13.56 & 23.36 & 
        34.68 & 13.60 & 22.76 \\
      321 & 61  &  (5.1\%) &
        {\bf 27.10} & {\bf 6.74} & {\bf 18.15} &
        33.36 & 14.36 & 23.53 &
        32.16 & 10.67 & 21.85 \\ 
      \hline
  \end{tabular}
  }
  \caption{Evaluation of pairwise alignment.
      Leftmost column indicates alignment between each sentence of the system summary and the oracle summary.
      The columns of `1st', `2nd' and `3rd' list the score of each sentence in the system summary.
      The \textbf{bolded} scores show the lowest scores.
      }
  \label{align_result}
\end{table*}

    \begin{table*}[t]
  \center
  \begin{subtable}{\textwidth}
  \scalebox{0.95}{
    \begin{tabular}{c|rrr|rrr|rrr|rrr}
      \hline
      & \multicolumn{3}{c|}{Coverage} & \multicolumn{3}{c|}{Parallel\_model}
        & \multicolumn{3}{c|}{Sequence\_model} & \multicolumn{3}{c}{Proposed} \\ \hline 
      & \multicolumn{1}{c}{R-1} & \multicolumn{1}{c}{R-2} & \multicolumn{1}{c|}{R-L} &
        \multicolumn{1}{c}{R-1} & \multicolumn{1}{c}{R-2} & \multicolumn{1}{c|}{R-L} &
        \multicolumn{1}{c}{R-1} & \multicolumn{1}{c}{R-2} & \multicolumn{1}{c|}{R-L} &
        \multicolumn{1}{c}{R-1} & \multicolumn{1}{c}{R-2} & \multicolumn{1}{c}{R-L} \\ \hline

      1st & 48.23 & 28.42 & 35.29 & 
            48.86 & 28.80 & 35.53 &
            49.20 & 28.84 & 35.38 &
            {\bf 49.48} & {\bf 29.15} & {\bf 35.82} \\
      2nd & 34.37 & 14.50 & 23.38 &
            34.30 & 14.18 & 23.28 & 
            {\bf 35.09} & {\bf 15.41} & {\bf 24.15} & 
            34.69 & 15.24 & 23.78 \\
      3rd & 31.91 & 12.36 & 22.00 & 
            {\bf 32.28} & {\bf 12.80} & {\bf 22.51} &
            32.01 & 12.18 & 21.97 &
            31.67 & 11.46 & 21.80\\
      Ave & 38.17 & 18.43 & 26.89 &
            38.48 & 18.59 & 27.11 &
            {\bf 38.77} & {\bf 18.81} & {\bf 27.17} &
            38.61 & 18.62 & 27.13\\
      \hline
    \end{tabular}
  }
  \subcaption{ Evaluation results in all test data.}  
  \label{result_all}
  \end{subtable}

  \center
  \begin{subtable}{\textwidth}
  \scalebox{0.95}{
    \begin{tabular}{c|rrr|rrr|rrr|rrr}
      \hline
      & \multicolumn{3}{c|}{Coverage} & \multicolumn{3}{c|}{Parallel\_model}
        & \multicolumn{3}{c|}{Sequence\_model} & \multicolumn{3}{c}{Proposed} \\ \hline 
      & \multicolumn{1}{c}{R-1} & \multicolumn{1}{c}{R-2} & \multicolumn{1}{c|}{R-L} &
        \multicolumn{1}{c}{R-1} & \multicolumn{1}{c}{R-2} & \multicolumn{1}{c|}{R-L} &
        \multicolumn{1}{c}{R-1} & \multicolumn{1}{c}{R-2} & \multicolumn{1}{c|}{R-L} &
        \multicolumn{1}{c}{R-1} & \multicolumn{1}{c}{R-2} & \multicolumn{1}{c}{R-L} \\ \hline
      1st & 48.38 & 28.97 & 35.68 &
            48.75 & 29.30 & 35.68 &
            49.35 & {\bf 29.38} & 35.68 &
            {\bf 49.63} & 29.18 & {\bf 35.95} \\
      2nd & 34.48 & 14.93 & 23.40 &
            33.62 & 13.89 & 22.81 & 
            34.32 &  14.97 &  23.64 &
            {\bf 34.51} & {\bf 15.13} & {\bf 23.74} \\
      3rd & 31.62 & 12.53 & 21.82 &
            {\bf 32.45} & {\bf 13.29} & {\bf 22.67} &
            31.58 & 12.50 & 21.80 &
            32.12 & 11.80 & 22.16 \\
      Ave & 38.16 & 18.81 & 26.97 &
            38.27 & 18.82 & 27.05 &
            38.42 & {\bf 18.95} & 27.04 &
            {\bf 38.75} & 18.70 & {\bf 27.28} \\
      \hline
    \end{tabular}
  }
  \subcaption{ Evaluation results in {\it{parallel}} test data.}  
  \label{result_par}
  \end{subtable}

  \center
  \begin{subtable}{\textwidth}
  \scalebox{0.95}{
    \begin{tabular}{c|rrr|rrr|rrr|rrr}
      \hline
      & \multicolumn{3}{c|}{Coverage} & \multicolumn{3}{c|}{Parallel\_model}
        & \multicolumn{3}{c|}{Sequence\_model} & \multicolumn{3}{c}{Proposed} \\ \hline 
      & \multicolumn{1}{c}{R-1} & \multicolumn{1}{c}{R-2} & \multicolumn{1}{c|}{R-L} &
        \multicolumn{1}{c}{R-1} & \multicolumn{1}{c}{R-2} & \multicolumn{1}{c|}{R-L} &
        \multicolumn{1}{c}{R-1} & \multicolumn{1}{c}{R-2} & \multicolumn{1}{c|}{R-L} &
        \multicolumn{1}{c}{R-1} & \multicolumn{1}{c}{R-2} & \multicolumn{1}{c}{R-L} \\ \hline

      1st & 47.83 & 26.92 & 34.25 &
            {\bf 49.18} & 27.45 & 35.14 &
            48.81 & 27.38 & 34.54 &
            49.09 & {\bf 29.08} & {\bf 35.47} \\
      2nd & 34.05 & 13.34 & 23.32 &
            36.14 & 14.96 & 24.55 &
            {\bf 37.18} & {\bf 16.58} & {\bf 25.53} &
            35.17 & 15.52 & 23.90 \\
      3rd & 32.68 & {\bf 11.93} & {\bf 22.47} &
            31.82 & 11.48 & 22.08 &
            {\bf 33.17} & 11.34 & 22.42 &
            30.46 & 10.55 & 20.84 \\
      ave & 38.19 & 17.40 & 26.68 &
            39.05 & 17.96 & 27.26 &
            {\bf 39.72} & {\bf 18.43} & {\bf 27.50} &
            38.24 & 18.38 & 26.74 \\
      \hline
    \end{tabular}
  }
  \subcaption{ Evaluation results in {\it{sequence}} test data.}  
  \label{result_seq}
  \end{subtable}

  \caption{Evaluation results per sentence breakdown.
    The scores in the row of `1st' are computed between the first sentence of the system summary and the sentence of the oracle summary.
    The rows of `2nd' and `3rd' are worked out in the same process. 
    The row of `ave' is the average scores of `1st', `2nd' and `3rd'.
  }
  \label{result_description}
\end{table*}

    \subsection{Experiment 1: Classification} \label{exp_cls:setup}
      \paragraph{Setup.}
      The articles and summaries were segmented using MeCab v0.996
      \footnote{\url{https://github.com/taku910/mecab}} (ipadic v2.7.0).
      The hidden states are represented as 256-dimensional matrices, while the
    word embeddings are 256-dimensional vectors.
      The size of the vocabulary is 2,350, which includes words that appear
      more than once.
      The model is trained using Adagrad \cite{Duchi:2011} with a learning rate of 0.01.

      The annotated portion of the development data is divided into 1,080 and
    120 training and test pairs, respectively.
      Because the summary structure is biased, as indicated by the results
      in Table \ref{anno_result}, we
      optimize our model by under-sampling to achieve high precision.
      In particular, we sample data for each label until its precision exceeds 0.8.

    \paragraph{Results using summary as input.} \label{exp_cls:result}
      Table \ref{cls_result} lists the test results and number of classified
    summaries for training data.
      We obtained 53,809 and 7,813 instances for automatically labeled
    parallel and sequence summaries, respectively.

    \paragraph{Results using article as input.}    \label{exp_cls:result2}
    Table \ref{cls_result2} lists the results of classification of the summary
    structures using articles as input.
    The accuracy of the binary classification is 0.50.
    Because the number of instances of sequence data is less than parallel
    data, its recall is lower than that of the parallel data.

  \subsection{Experiment 2: Summarization} \label{exp_sum}
    \paragraph{Setup.}
    In our experiment, we use two models: the baseline \cite{See:17} and proposed models.
    Similar to \cite{See:17}, the hidden states of these two models are
    represented by 256-dimensional matrices, while the word embeddings are
    represented by 128-dimensional vectors.
    The vocabulary size is 50,000 words in the case of both the source and
    target documents.

    Furthermore, we also followed \newcite{See:17} to perform several
data cleanup operations: the articles were truncated to 400 words from the beginning of
the article for the training and test
    sets, while the summaries shorter than 70 words were excluded.
    The models were trained using Adagrad \cite{Duchi:2011} with a learning
rate of 0.15, and gradient clipping with a maximum gradient norm of 2.

  \paragraph{Evaluation metric.} \label{eval_metric}
    We use the F$_1$ scores for ROUGE-1, ROUGE-2 and ROUGE-L \cite{Lin:04} to
    evaluate the output;
    this evaluation is applied to all test cases, {\it{parallel}}
    summaries, and {\it{sequence}} summaries, respectively.

    Furthermore, we align each sentence in the system summary with a
    sentence of an oracle summary to evaluate consistency.
    The alignment is selected to maximize the average score of
    ROUGE-L under the condition that there is no duplicate.

  \paragraph{Result.} \label{chap:result}
    Tables \ref{plain_result} and \ref{result_description} list the
evaluation results for all, parallel, and sequence test data.
    As can be seen from Tables \ref{plain_result} and \ref{result_all}, the proposed models
outperform the baseline model (Coverage).
%
    It is clear from Table \ref{result_par}, in a manner similar to the
    results in Table \ref{result_all}, the sequence model outperforms
    the others on average.
    However, as is evident from Table \ref{result_seq}, each model behaves differently.
    We explore the reason for this in the next section.

  \section{Discussion} \label{discuss}

    \subsection{Model Comparison} \label{discuss:result}
    For the `All' values in Table \ref{plain_result} and `Ave' values in Table
    \ref{result_all}, the scores of the sequence model are the highest.
    It can be said that the sequence model considers previous sentences more
    than the parallel model while generating the third sentence in a summary.
    The parallel model fails to incorporate the information of the second
    sentence while generating the third sentence (Table \ref{result_seq}).

    Table \ref{tab:example} shows an example of summarization.
    The example was annotated as a parallel type. The first sentence
    in the reference mentions the result of a game. However, it is difficult
    for all models to generate this information because it is not described in the article.
    Among all the models, only the sequence model produces a sentence similar to
    the reference. This can be attributed to its ability to generate
    the second sentence based on previous information more effectively than the others models.


    \subsection{Pairwise Evaluation} \label{discuss:EEP}
      To analyze the system summary in greater detail, we evaluate the
      performance of our proposed method for each pair combination.
      Table \ref{align_result} lists the results of our pairwise evaluation.
    
      The pairs of `123' occurred most frequently.
      It is considered that the summaries in the case of sequence or parallel
    types are suitably generated.
      These scores for the pairs of `123' are higher than the others.
      The pairs of `132' account for the second largest proportion of pair
    combinations; their scores are almost as high as those of the pairs of `123'.
      This is because the second and third sentences in a parallel type
    summary are in no particular order.
    
      In contrast, the score of the sentences in the reverse order is low
      (e.g., the scores shown in bold in the row of `213' and the
    second column).
      Thus, based on the above discussion, it is suggested that the first sentence
      should be correctly generated in order to evaluate the summary
      structure appropriately.


    \subsection{Evaluation Methods} \label{discuss:methods}
      In this work, we evaluated the generated summaries using two methods of evaluation.
      We applied the ordinary evaluation, which is discussed in Section
    \ref{chap:result}, as well as the evaluation of each pair based on the
ROUGE-L score,
    which is discussed in Section \ref{discuss:EEP}.

      The ordinary evaluation indicates summary informativeness.
      In contrast, the evaluation of each pair shows how the proposed model
    generated summary sentences.
      It should be noted that the generated summary does not consist of the
    same order of sentence compared with the oracle summary.
      By performing pairwise evaluation using ROUGE-L, we were able to
      evaluate the order of the summary appropriately for the dataset.

  \section{Conclusion} \label{conc}
    In this study, we constructed a dataset focused on summaries with three sentences.
    We annotated and analyzed the structure of the summaries in the
considered dataset.
    In particular, we proposed a structure-aware summarization model combining
    the summary structure classification model and summary-specific summarization
    sub-models.
    Through our experiment, we demonstrated that our proposed model
    improves summarization performance over the baseline model.

    In future work, we will use category and subcategory tags to analyze
    the characteristics for each category so that we can build specific models
    to improve the summarization system.

  \bibliography{paclic32}
  \bibliographystyle{acl}

\section*{Appendix A. Example of Generated Summary}

    \begin{table*}
\begin{tabular}{p{\linewidth}}
\toprule
\textit{Source} \\
Cologne's local paper Express praised that Yuya Osako demonstrated its power
in the game against Wolfsburg the other day. It is reported that he
demonstrated his power as a Japan representative FW who
showed up trying to toss the Wolfsburg defense team.\\
In the last season, due to insufficient scoring ability, he
received fierce criticism from passionate Cologne fans so that Cologne
coach St\"{o}ger dare to avoid appointment in a home game.\\
Nonetheless head coach St\"{o}ger and manager Joerg Schmadtke is constantly
defending Osako as ``a wonderful football player.''\\
On the other hand, CF Modeste is showing off his team's top scorers, ``I knew
that his main job was CF,'' said St\"{o}ger, ``but I really wanted to appoint
him,'' resulting in various positions by trial and error.\\
But this season's Cologne has adopted the 4-4-2 system from the opening, ``I am
clear now. He will play as CF,'' he asserted. Osako got two goals in
the first round of Pokal, and also scored in the test match during
international matches.
\\
And the first opportunity of this season, due to Rudnevs' injury,
Osako showed off a success that the Express paper praised as ``Osako
showed ball-keeping ability, speed, breadth of vision, and scoring ability
through the game.'' In Cologne's FW team, he was even appreciated as the
only player who achieves all the performance.\\
As for the score, this day he did not score at all because Casteels played
a blinder, but he wants to maintain the good performance at CF as his professional career so as to secure a fixed position. \\
\midrule
\textit{Reference} \\
Cologne to which Yuya Osako belongs drew 0-0 in Wolfsburg on the 10th \\
A local newspaper praised that Osako tossed at the Wolfsburg defense team \\
In Cologne's FW team, he was described as having all the performance \\
\midrule
\textit{Coverage}: ROUGE-L=47.02 \\
Cologne's local paper Express reported that Yuya Osako demonstrated his power \\
It defends Osako as ``a wonderful football player'' in the 2015 Wolfsburg game
\\
``I understood that I am clear now,'' he said \\
\midrule
\textit{Parallel\_model}: ROUGE-L=47.16 \\
Cologne's local paper Express recieved a fierece criticism from the team top
scorers fan \\
``I knew that CF was his main job," said head coach St\"{o}ger with a blinder \\
Last season, Yuya Osako said ``I am clear now, he will play as CF'' \\
\midrule
\textit{Sequence\_model}: ROUGE-1=47.43 \\
Cologne's local paper Express reported that Yuya Osako demonstrated his power
in the Wolfsburg game \\
Japan representative FW who tried to toss at the Wolfsburg defense team,
is praised as he demonstrated his performance \\
``I understood that I am clear now,'' as he defended \\
\bottomrule
\end{tabular}
\caption{Example of generated three bullet points summarization in
\textit{parallel} structure.
Retrieved from \url{http://news.livedoor.com/article/detail/12016209/}
on the 11th January, 2018.}
\label{tab:example}
\end{table*}

\end{document}